\documentclass{article}
\usepackage[latin9]{inputenc}
\usepackage{verbatim}
\usepackage{float}
\usepackage{amsmath}
\usepackage{amssymb}
\usepackage{graphicx}

\makeatletter

\providecommand{\tabularnewline}{\\}
\floatstyle{ruled}
\newfloat{algorithm}{tbp}{loa}
\providecommand{\algorithmname}{Algorithm}
\floatname{algorithm}{\protect\algorithmname}

\usepackage{spconf,amsmath,graphicx}

\usepackage{times}
\usepackage{cite}
\usepackage{amsfonts}
\usepackage{mathrsfs}
\usepackage{algorithm}
\usepackage{algorithmic}
\usepackage{epsfig}
\usepackage[amsmath,hyperref,thmmarks]{ntheorem}

\newtheorem{theorem}{Theorem}

\makeatother

\title{Multi-view Metric Learning for Multi-view Video Summarization}

\name{Yanwei Fu$^{\dagger}$, Lingbo Wang$^{\wedge}$, Yanwen Guo$^{\star}$ }
\address{$^{\dagger}$ Disney Research Pittsburgh \\
$^{\wedge}$ the Key Laboratory of Intelligent Computing and Signal
Processing, \\School of Computer Science and Technology, Ministry
of Education, Anhui University, China  \\
$^{\star}$ the National Key Laboratory for Novel Software Technology,
Nanjing University, China\\
}

\begin{document}
\maketitle

\begin{abstract}
Traditional methods on video summarization are designed to generate
summaries for single-view video records; and thus they cannot fully exploit
the redundancy in multi-view video records. In this paper, we present
a multi-view metric learning framework for multi-view video summarization
that combines the advantages of maximum margin clustering with the
disagreement minimization criterion. The learning framework thus has
the ability to find a metric that best separates the data, and meanwhile
to force the learned metric to maintain original intrinsic information
between data points, for example geometric information. Facilitated
by such a framework, a systematic solution to the multi-view video
summarization problem is developed. To the best of our knowledge,
it is  the first time to address multi-view video summarization
from the viewpoint of metric learning. The effectiveness of the proposed
method is demonstrated by experiments.
\end{abstract}

\section{Introduction}

In many real world applications, unlabeled data usually arrive in
the form of a number of highly correlated views. Examples of this
kind can be frequently encountered in the field of video processing,
where different cameras may focus on roughly the same field-of-view
(Fov) from different viewpoints, such as in the case of office coverage
or surveillance records. In such case, one might expect to utilize
correlations to help understand and characterize the data, and more
preferably, find an ``optimal'' metric that reflects the intrinsic
structure of the input data.

In this paper, we are interested in this kind of problem, especially
multi-view video summarization. Suppose the complicated human motion
in local geometric coordination is a function varying with time and
sampled temporally by multiple cameras simultaneously. In order to
reveal the characteristics of this original space, traditional methods
generally extract high dimensional feature vector space for each view
video with a manifold assumption individually. Many dimension reduction
methods are then utilized. However, different view videos often include
distinctive and complementary information to the original dataset.
For this purpose, we present multi-view metric learning framework
to integrate all views of these videos into the new metric learning
space and to disclose the intrinsic features of original human moving.
Here, video summarization is such intrinsic feature we are striving
for.

\begin{figure*}[htp]
\centering{}\includegraphics[scale=0.4]{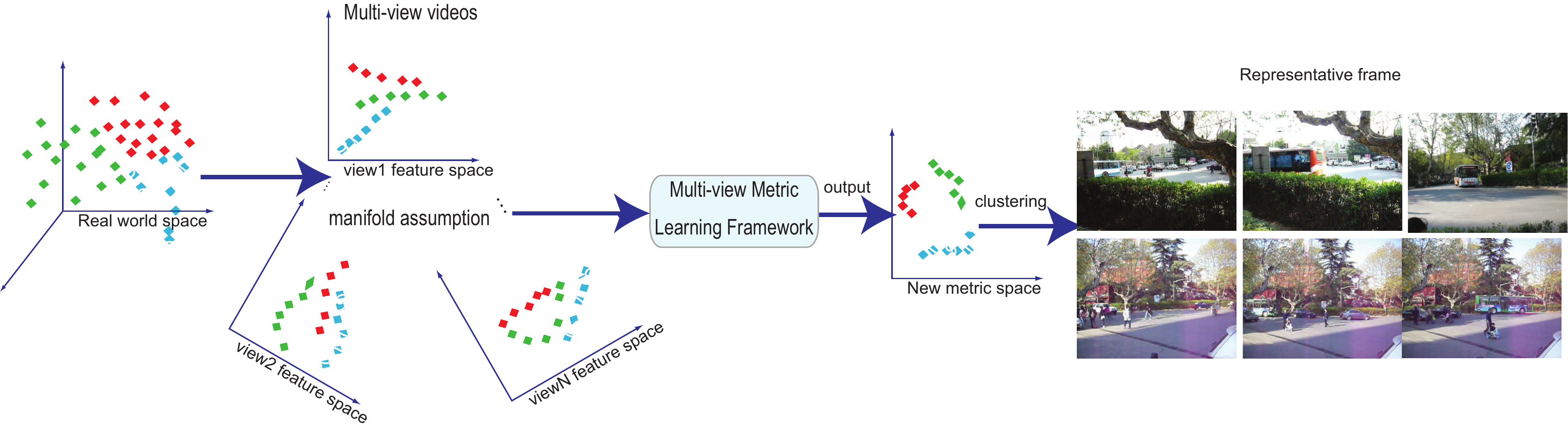}\caption{{\small \label{fig:The-flowchar-of}The flowchar of the multi-view
metric learning framework with application to multi-view video summarization. }}
\end{figure*}

We firstly provide a unified framework for multi-view video summarization
by multi-view metric learning. Multi-view video simultaneously captures
the different visual projections of the same time-space manifold in
real life. Our multi-view metric learning is learned to project all
the multi-view videos into a new metric space to best simulate the
real world manifold space. This thus greatly facilitates the video
summarization by preserving most intrinsic features across different
views. Specifically, the framework is derived from Maximal Margin
Clustering (MMC) by minimizing the disagreement minimization criterion
for learned metric. In the learned metric space, visual data are summarized
by clustering them and extracting key frames in each cluster.

\section{Previous work}

Multi-view learning has received considerable attention in the past
decade. Most of previous methods are however devoted to the semi-supervised
learning (\cite{Zhu05Survey} for a detailed survey). Some studies
on the unsupervised case have been performed \cite{daSa05Spectral,Zhou07MultipleView,Long08GeneralModel},
focusing on merging different metrics and minimizing disagreements
among them. Our approach is different with them in that we take simultaneously
the Minimizing-Disagreement criterion and the Maximal Margin Clustering
(MMC) criterion into account. The optimization involved achieves a
trade-off between them.

Maximum Margin Clustering (MMC) is a classical approach to clustering\cite{Xu05MMC,Xu05Unsupervised}
that aims at finding clusters with large margins. It often exhibits
superior performance compared against traditional clustering algorithms.
Following the same criteria as MMC approaches, the method proposed
in this paper optimizes a graph-theoretic measure to find a kernel
matrix that allows larger margins between clusters.

Metric learning \cite{Xing03Distance,Weinberger06Metric} and multiple
kernel learning\cite{Lanckriet04MultiKernel} aim at finding an ``optimal''
distance metric (or convex combination of kernels that implicitly
defines a distance metric) that allows distance-based or kernel-based
algorithms to achieve better performance. Previous studies on metric
learning and multiple kernel learning mainly focus on situations where
additional information (such as side information \cite{Xing03Distance}
or class labels\cite{Weinberger06Metric,Zien07MulticlaskMKL,Argyriou06Combine})
is available. Studies on ``pure'' unsupervised learning only utilize
either the maximum margin criteria\cite{Yeh08MKC,Zhao09MKC} or the
minimizing-disagreement criteria \cite{Long08GeneralModel}.

Video summarization is a well studied topic in the past two decades.
We refer \cite{Truong07Video} for a comprehensive survey. Although
some previous studies have been dedicated to the problem of multi-camera
systems, but they were either focusing on tracking moving objects
across cameras with non-overlapping field of views \cite{Prosser08MCMatching:BMVC,Loy09MCActivity}
or compression \cite{Vetro07MVCOverview}. Fu \emph{et al}. \cite{Fu2010}
wass the first effort to systematically study the problem of skim-based multi-view
video summarization (especially in the surveillance videos) by using
hypergraph structures. \cite{pingli11ICIP}  extracted the keyframes of such multi-view summarization.
 We explore this problem by multi-view metric learning
framework. We instead directly address the problem of video summarization
on multiple overlapping views.

\section{Multi-view metric learning framework}

\label{sec:framework}

Suppose $\left\{ X^{(k)}\right\} _{k=1}^{K}$ is the the low-level
features of $K$ different views, where $X^{(k)}=[\mathbf{x}_{1}^{k}\dots\mathbf{x}_{n}^{(k)}]'\in R^{n\times d_{i}}$
are the coordinate matrices. Our goal is to find a unified coordinate
matrix $X=\left[\mathbf{x}_{1}\dots\mathbf{x}_{n}\right]^{T}\in R^{n\times d}$
minimizing
\begin{equation}
R(X)=R_{\text{emp}}(X)+\gamma_{1}R_{\text{struct}}(X)+\gamma_{2}R_{\text{diff}}(X)\label{eqn:R}
\end{equation}
 where $R_{\text{emp}}(X),R_{\text{struct}}(X),R_{\text{diff}}(X)$
are the empirical, structural, and disagreement losses of $X$, respectively.
$\gamma_{1},\gamma_{2}$ are parameters controlling the trade-off
of objectives.

The classical \textbf{MMC} contains the former two parts: $R_{\text{emp}}(X),R_{\text{struct}}(X)$.
However, this problem requires the new metric learning must preserve
some important information of data points in original space. Therefore,
disagreement minimization criterion (\textbf{DMC}) is added by $R_{\text{diff}}(X)$.

The empirical loss $R_{\text{emp}}(X)$ is usually defined according
to label information (such as labels of instances or certain ``side
information''). For example, in supervised multiple kernel learning,
$R_{\text{emp}}(X)$ is usually defined as the minimum hinge loss
achievable on the metric defined by $X$. The structural loss $R_{\text{struct}}(X)$
can be defined as complexity of classifiers (as in the case of SVM),
or be used to ensure ``similar'' instances have ``similar'' labels
(as in the some formulation of manifold learning, e.g. \cite{Belkin06ManifoldRegularization}),
etc. The disagreement loss $R_{\text{diff}}(X)$ measures the extent
to which $X$ is different with the $X^{(k)}$.

\subsection{ Unsupervised multi-view metric learning}

This section discusses the choice of each loss function for the framework.

First, suppose $G^{(1)},\dots,G^{(K)}\in R^{n\times n}$ are the similarity
matrices defined by the metric spaces $X^{(1)},\dots,X^{(K)}$, respectively,
where $G^{(k)}(i,j)=G^{(k)}(j,i)$ is the similarity between data
points $x_{i}$ and $x_{j}$ on the $k$-th view%
\footnote{We use RBF kernel $G(i,j)=\exp(-\frac{\parallel\mathbf{x}_{i}-\mathbf{x}_{j}\parallel^{2}}{2\sigma^{2}})$
to define similarity. \cite{Luxburg07Tutorial} has a full discussion.%
}. Let $\mathcal{L}^{(k)}$ be the normalized Laplacian of $G^{(k)}$,
where the normalized Laplacian of a similarity matrix $G$ is defined
as
\begin{equation}
\mathcal{L}(G)=I-D_{G}^{-1/2}GD_{G}^{-1/2}
\end{equation}
 where $D_{G}=\Lambda\{\deg_{G}(1),\dots,\deg_{G}(K)\}$ with $\deg_{G}(i)=\sum_{j=1}^{n}G(i,j)$.  And I
  is the eye matrix.

A good video summary will have a better coordinated $R_{\text{diff}}(X)$
and invariant to the metric transformations of synchronous frames,
such as rotation, translation, and scaling. More subtly, it is nontrivial
to make the framework robust to different visual conditions especially
for surveillance video summary. To this end, we define the disagreement
loss as
\begin{equation}
R_{\text{diff}}(X)=\sum_{k=1}^{K}\parallel\frac{1}{tr(\mathcal{L}(G_{X}))}\mathcal{L}(G_{X})-\frac{1}{tr(\mathcal{L}^{(k)})}\mathcal{L}^{(k)}\parallel_{F}^{2}\label{eqn:R:diff:laplacian}
\end{equation}

$G_{X}$ is the similarity transformation of the metric $X$.
This function can be viewed as a simplified version of the Canonical
Correlation Analysis (CCA) \cite{Hardoon04CCA} measure. Like CCA,
it is invariant to certain kinds of metric transformations such as
rotation, translation, and scaling and better coordinate different
visual conditions. Furthermore, it is more desirable in that it introduces
no optimization variables.

Our definition of $R_{\text{struct}}(X)$ is motivated by the following
results on spectral graph theory.

\begin{theorem}[\cite{Luxburg07Tutorial}] The multiplicity $c$ of
the eigenvalue $0$ of $\mathcal{L}(G)$ equals the number of connected
components $A_{1},\dots A_{c}$ in the graph. \end{theorem}

\begin{theorem}[\cite{Chung97Graph}] For $S_{i}\subset V(G),i=0,1,\dots c$,
we have
\[
\min_{i\neq j}d(S_{i},S_{j})\leq\max\lceil\frac{\log\sqrt{\frac{\mathrm{vol\,}\bar{S}_{i}\mathrm{vol\,}\bar{S}_{j}}{\mathrm{vol\,}S_{i}\mathrm{vol\,}S_{j}}}}{\log\frac{\lambda_{n-1}+\lambda_{c}}{\lambda_{n-1}-\lambda_{c}}}\rceil
\]
 where $d(S_{i},S_{j})=\min\left\{ d(x_{s},x_{t}):x_{s}\in S_{i},x_{t}\in S_{j}\right\} $,
$d(x_{s},x_{t})$ is the shortest path from $x_{s}$ to $x_{t}$,
$\mathrm{vol}\, S_{i}=\sum_{x_{j}\in S_{i}}\deg(j)$, $\bar{S_{i}}=V(G)-S_{i}$,
and $\lambda_{0}\leq\lambda_{1}\leq\dots\leq\lambda_{n-1}$ are the
eigenvalues of $\mathcal{L}(G)$. \end{theorem}

These theorems indicate that the first $k$ smallest eigenvalues of
$\mathcal{L}(G_{X})$ determine the quality of $k$-clustering on
the metric implicitly defined by $G_{X}$ (which is a transformation
of the metric $X$). Therefore, we define the structural loss as
\begin{equation}
R_{\text{struct}}(X)=\frac{\sum_{i=0}^{c-1}\lambda_{i}}{\sum_{i=0}^{n-1}\lambda_{i}}=\frac{1}{tr(\mathcal{L}(G_{X}))}\sum_{i=0}^{c-1}\lambda_{i}\label{eqn:R:struct:L}
\end{equation}
 where $\lambda_{0}\leq\lambda_{1}\leq\dots\leq\lambda_{n-1}$ are
the eigenvalues of $\mathcal{L}(G_{X})$, and $c$ is a parameter
indicating the desired number of clusters.

Finally, the unsupervised learning settings donot have label information,
we simply let $R_{\text{emp}}(X)\equiv0$

Combining the definitions above, we finally formulate our optimization
objective for unsupervised multi-view metric learning as
\begin{align}
\min_{X}R(X) & =\frac{1}{tr(\mathcal{L}(G_{X}))}\sum_{i=0}^{c-1}\lambda_{i}\label{eqn:R:final}\\
+ & \gamma\sum_{k=1}^{K}\parallel\frac{1}{tr(\mathcal{L}(G_{X}))}\mathcal{L}(G_{X})-\frac{1}{tr(\mathcal{L}^{(k)})}\mathcal{L}^{(k)}\parallel_{F}^{2}
\end{align}

\subsection{Discussion for some alternative choices}

\label{sec:discussion}

As $R_{\text{diff}}(X)$ is the measure of disagreement between metric
spaces, one may consider the CCA as a good choice. However, the calculation
of CCA involves optimization on transformation matrices, which will
introduce $\Omega(n^{2})$ optimization variables into the optimization
problem, making the optimization intractable.

A simplification of the CCA measure leads to the following prediction-based
disagreement measure \cite{Wang08MultiKMHKS}:
\begin{equation}
R_{\text{diff}}(X)=\sum_{k=1}^{K}\parallel f_{X}-f^{(k)}\parallel^{2}\label{eqn:R:diff:result}
\end{equation}
 where $f_{X}$ and $f^{(k)}\in R^{n}$ denote, the prediction of
the classifier learned according to the metrics $X$ and $X^{(k)}$.
This definition is advisable when classification results can be easily
deduced from the learned metric in the same optimization framework.
Yet problems arise when we are facing clustering tasks, where the
disagreement between different clustering results may be difficult
to calculate.

Compared with these definitions, our definition of disagreement loss
is more straightforward and computationally efficient as it is directly
based on the metric learned and introduces no additional optimization
variables.

\begin{comment}

\subsection{Relationship with multiple kernel learning}

Although the form of Algorithm~\ref{alg:simple} is similar to that
of multiple kernel learning \cite{Lanckriet04MultiKernel}, the multi-view
metric learning is different to multiple kernel learning.
\begin{itemize}
\item Multiple kernel learning treats kernels as transformations of metrics
and does not try to capture the semantic meaning of the kernels. So
it does not enforce an agreement between learned kernel and original
ones. On the contrary, in multi-view metric learning, the semantic
meaning captured by each view is preserved by the introduction of
disagreement loss.
\item Lacking disagreement loss, the multiple kernel learning must be guided
by measure of certain explicit classification result, which makes
it \end{itemize}
\end{comment}

\section{Optimization for unsupervised multi-view metric learning}

\label{sec:algorithm}

In this section, we present an efficient algorithm for solving the
optimization problem in Eq.~\ref{eqn:R:final}.

Let $\mathcal{\hat{L}}=\frac{1}{tr(\mathcal{L}(G_{X}))}\mathcal{L}(G_{X})$
and $\mathcal{\hat{L}}^{(k)}=\frac{1}{tr(\mathcal{L}^{(k)})}\mathcal{L}^{(k)}$.
And note that, once the $\mathcal{\hat{L}}$ is found, a metric space
is implicitly defined. In fact, given $\mathcal{\hat{L}}$, the coordinate
matrix $X_{c}=[\Phi_{1}\dots\Phi_{c}]$ is a metric space, where $\Phi_{i}$
is the eigenvector corresponding to the $i$-th smallest eigenvalue
of $\mathcal{\hat{L}}$, and $k$-means algorithm can be used for
clustering according to this metric space. This is exactly the way
in which normalized cut on a graph $G$ is usually performed \cite{Luxburg07Tutorial}.
Therefore, for the purpose of clustering, it suffices to compute the
$\mathcal{\hat{L}}$ itself (note that, $\mathcal{\hat{L}}$ has the
same eigenvectors as $\mathcal{L}(G_{X})$ and therefore leads to
the same clustering result). The optimization problem now turns to
\begin{equation}
\min_{\mathcal{\hat{L}}\preceq\mathbf{1},tr(\mathcal{\hat{L}})=1}\sum_{i=0}^{c-1}\lambda_{i}+\gamma\sum_{l=1}^{m}\parallel\mathcal{\hat{L}}-\mathcal{\hat{L}}_{l}\parallel_{F}^{2}\label{eqn:opt:goal:1}
\end{equation}

With consideration of efficiency, we further assume that $\mathcal{\hat{L}}=\sum_{k=1}^{K}\mu_{k}\mathcal{\hat{L}}^{(k)}$.
It can be efficiently solved by alternating descent method: firstly
fixed $\mathcal{\hat{L}}$, $P$ can be solved via eigen-decomposition
of $\mathcal{\hat{L}}$; then fixed $P$, $\mu$ is solved by a quadratic
programming (Eq.\ref{eqn:small_qp}) until convergence. This quadratic
programming problem can be efficiently solved by Mosek in that $m$
is always small in practice $m\sim\mathcal{O}(1)$:
\begin{equation}
\min_{\mu\geq0,\sum_{k=1}^{K}\mu_{k}=1}\sum_{k=1}^{K}\mu_{k}tr(P'\mathcal{\hat{L}}^{(k)}P)+\sum_{i=1}^{K}\parallel\sum_{k=1}^{K}\mu_{k}\mathcal{\hat{L}}^{(k)}-\mathcal{\hat{L}}^{(i)}\parallel_{F}^{2}\label{eqn:small_qp}
\end{equation}

\section{Application to multi-view video summarization\label{sec:video}}

To generate video summary, we assume that each event $E_{i}$ in the
real world corresponds to a distribution $D_{i}$ centered at a small
region in a ``latent'' semantic space. Each ``instance'' of the
event $E_{i}$ is a data point $x_{ij}$ sampled according to $D_{i}$
in the latent semantic space.

Our solution to multi-view video summarization is summarized in Algorithm~\ref{alg:overall}.We
deal with videos of the same spot with different angles, so the high-dimensional
low-level features of each view is embedded in the same low-dimensional
space. This justifies the usage of the above-mentioned framework,
which imposes a disagreement-minimization criteria on the metric learning.

\begin{algorithm}[htbp]
\caption{Multi-view video summarization.\label{alg:overall}}

\begin{enumerate}
\item Decompose video records into sets of frames, denoting as $X^{(1)},\cdots,X^{(K)}$,
where $X^{(k)}=[\mathbf{x}_{1}^{(k)}\dots\mathbf{x}_{n}^{(k)}]'\in R^{n\times d_{k}}$
is the $d_{k}$-dimensional feature representation of the $n$ frames
in the $k$-th view.
\item Learn a unified metric space $X\in R^{n\times d}$ according to the
information lying in $X^{(1)},\cdots,X^{(K)}$.
\item Perform clustering on $X$, using the centers of clusters as representatives,
denoting as $F=\left\{ f_{i_{1}},\cdots,f_{i_{C}}:1\leq i_{1}<\cdots<i_{C}\leq n\right\} $.
\item Select a frame for each $f_{i_{c}}$ out of the $K$ frames corresponding
to it, and output these frames as the final summary.\end{enumerate}
\end{algorithm}

\section{Experiments}

\label{sec:experiment}

We conduct our experiments on Road and Office1 datasets \cite{Fu2010}
which is captured by three hand-held video cameras with 360 degree
coverage of the scene. Some representative frames are shown in Fig.
\ref{fig:The-flowchar-of}. The same important objects (bus or human)
are highlighted and extracted from original different views. This
facilitates the quick browsing and understanding the original videos
with overlapping views. For baselines, we construct a graph for the
frames in each view, employ normalized cut for clustering and select
the representative frames. ED (Euclidean distance) method utilizes original feature vector
space (Euclidean space) of each view for metric learning, while DM
methods use Diffusion metric for metric learning.

We employ the groundtruth of important events of Office1 dataset defined
in \cite{Fu2010} to measure the objectiveness performance. We reported
the results in \cite{Fu2010} and extract the same length summary
for Uni., Ran., ED, DM and our method in Tab.\ref{tab:Objectively-performance-comparis}.
The results shows that our method is better than the other methods.

\begin{table}
\begin{centering}
{\small }%
\begin{tabular}{|c|c|c|c|}
\hline
{\small Methods} & {\small No. Eve.} & {\small Precision($\%$)} & {\small Recall($\%$)}\tabularnewline
\hline
\hline
{\small \cite{Fu2010}} & {\small 16} & {\small 100} & {\small 61}\tabularnewline
\hline
{\small Uni./Ran.} & {\small 10/5} & {\small 70/60} & {\small 26.9/11.5}\tabularnewline
\hline
ED/DM & 10/13 & 80/76.9 & 30.8/38.5\tabularnewline
\hline
{\small Ours} & {\small 20} & {\small 100} & {\small 76.9}\tabularnewline
\hline
\end{tabular}
\par\end{centering}{\small \par}

\begin{centering}
{\small }
\par\end{centering}{\small \par}

\caption{\label{tab:Objectively-performance-comparis}Objectively performance
comparison with previous methods on office1. Uni. means we uniformly
summarize the videos, while Ran. indicates we randomly summarize the
frames of videos.}

\end{table}

To further evaluate the effectiveness of these five methods, we conduct
user study by inviting 12 participants and gave their judgements for
the results. Table \ref{expe_table} shows the scores which are normalized
from 0 to 1 and higher scores indicate better satisfaction. The summary results of \cite{Fu2010} are
not directly comparable in this part. Because it is skim-based
summary while ours are keyframe summary. It shows that the learned
multi-view metric space can improve the user satisfaction than other
baselines.
\begin{table}[tp]
\begin{centering}
\caption{\label{expe_table}Statistical data of user study.}

\par\end{centering}

\centering{}%
\begin{tabular}{||p{1.63cm}|p{2.41cm}|p{2.56cm}||}
\hline
\multicolumn{1}{|p{1.63cm}|}{} & \multicolumn{1}{p{2.41cm}|}{ road} & \multicolumn{1}{p{2.56cm}|}{ office1}\tabularnewline
\hline
\multicolumn{1}{|p{1.63cm}|}{Uni./Ran.} & \multicolumn{1}{p{2.41cm}|}{ 0.4/0.3} & \multicolumn{1}{p{2.56cm}|}{ 0.45/0.35}\tabularnewline
\hline
\multicolumn{1}{|p{1.63cm}|}{ ED/DM} & \multicolumn{1}{p{2.41cm}|}{ 0.68/0.78} & \multicolumn{1}{p{2.56cm}|}{ 0.72/0.75}\tabularnewline
\hline
\multicolumn{1}{|p{1.63cm}|}{ Ours} & \multicolumn{1}{p{2.41cm}|}{ 0.80} & \multicolumn{1}{p{2.56cm}|}{ 0.76}\tabularnewline
\hline
\end{tabular}
\end{table}

\section{Conclusion}

\label{sec:conclusion}

In this paper, we present a systematic solution to multi-view video
summarization. The solution is based on reconstructing the latent
semantic metric by multi-view metric learning. The multi-view metric
learning method achieves a balance between the separability of clusters
and the similarity to original metrics with an efficient optimization
algorithm.

The multi-view metric learning algorithm proposed in the paper can
be used to efficiently learn an ``optimal'' combination of multiple
metrics. The ``optimality'' is defined as a trade-off between the
maximum margin between clusters achievable on the metric and the similarity
between the learned metric and the original ones.

%The proposed solution to multi-view video summarization finds representative
%nodes in the merged metric. It resorts to the original metric spaces
%for the final choice of keyframes. Further improvement on the choice
%of keyframes may be made by evaluating the contribution of each frame
%to the learned metric.

{\small \bibliographystyle{abbrv}
\bibliography{xiaoxp}
}
\end{document}